\documentclass[conference]{IEEEtran}
\IEEEoverridecommandlockouts
\usepackage{cite}
\usepackage{amsmath,amssymb,amsfonts}
\usepackage[numbers]{natbib}
\usepackage{algorithmic}
\usepackage{graphicx}
\usepackage{algorithm}
\usepackage{subfigure}
\usepackage{changes}
\usepackage{hyperref}
\hypersetup{hidelinks,
	colorlinks=true,
	allcolors=black,
	pdfstartview=Fit,
	breaklinks=true}
\def\BibTeX{{\rm B\kern-.05em{\sc i\kern-.025em b}\kern-.08em
    T\kern-.1667em\lower.7ex\hbox{E}\kern-.125emX}}
\begin{document}

\title{Balancing Exploration and Exploitation in Hierarchical Reinforcement Learning via Latent Landmark Graphs}

\author{\IEEEauthorblockN{Qingyang Zhang$^{1,2}$\thanks{This work is supported in part by National Key R\&D Program of China (No.2022ZD0116405), in part by the Program for National Nature Science Foundation of China (62073324), and in part by the Strategic Priority Research Program of the Chinese Academy of Sciences (No.XDA27030300).},
Yiming Yang$^{1}$,
Jingqing Ruan$^{1,2}$, 
Xuantang Xiong$^{1,3}$, 
Dengpeng Xing$^{1,3,*}$\thanks{*Corresponding author.}, and
Bo Xu$^{1,2,3}$}
\IEEEauthorblockA{\textsuperscript{\rm 1}Institute of Automation, Chinese Academy of Sciences, Beijing, China}
\IEEEauthorblockA{\textsuperscript{\rm 2}School of Future Technology, University of Chinese Academy of Sciences, Beijing, China}
\IEEEauthorblockA{\textsuperscript{\rm 3}School of Artificial Intelligence, University of Chinese Academy of Sciences, Beijing, China}
\tt \{zhangqingyang2019, yangyiming2019, ruanjingqing2019, xiongxuantang2021, \\ dengpeng.xing, xubo\}@ia.ac.cn
}

\maketitle

\begin{abstract}
Goal-Conditioned Hierarchical Reinforcement Learning (GCHRL) is a promising paradigm to address the exploration-exploitation dilemma in reinforcement learning. It decomposes the source task into subgoal conditional subtasks and conducts exploration and exploitation in the subgoal space. The effectiveness of GCHRL heavily relies on subgoal representation functions and subgoal selection strategy. However, existing works often overlook the temporal coherence in GCHRL when learning latent subgoal representations and lack an efficient subgoal selection strategy that balances exploration and exploitation. This paper proposes \textbf{HI}erarchical reinforcement learning via dynamically building \textbf{L}atent \textbf{L}andmark graphs (HILL) to overcome these limitations. HILL learns latent subgoal representations that satisfy temporal coherence using a contrastive representation learning objective.
Based on these representations, HILL dynamically builds latent landmark graphs and employs a novelty measure on nodes and a utility measure on edges. Finally, HILL develops a subgoal selection strategy that balances exploration and exploitation by jointly considering both measures. Experimental results demonstrate that HILL outperforms state-of-the-art baselines on continuous control tasks with sparse rewards in sample efficiency and asymptotic performance\footnote{Our paper has been accepted by the conference of International Joint Conference on Neural Networks (IJCNN) 2023.}.
Our code is available at~\href{https://github.com/papercode2022/HILL}{\textcolor{magenta}{https://github.com/papercode2022/HILL}}.
\end{abstract}


\section{Introduction}
Balancing exploration and exploitation is a one of the major challenges in reinforcement learning.
Goal-Conditioned Hierarchical Reinforcement Learning (GCHRL)~\cite{dayan1992feudal,sutton1999between} is a promising paradigm that leverages task decomposition and temporal abstraction to address this challenge.
GCHRL typically consists of two hierarchical levels~\cite{barto2003recent}, where a high-level controller periodically decomposes the source task into subgoal conditional subtasks, and a low-level controller learns to complete those subtasks by reaching the subgoals. The subgoal sequence helps compress the exploration space, thereby reducing the exploration difficulty. It also enables efficient exploitation by selecting subgoals with higher value estimations.

A proper subgoal representation function is crucial for effective GCHRL. Early efforts manually identify bottleneck states as subgoals \cite{kulkarni2016hierarchical,nachum2018data}, which require task-specific knowledge. Recent approaches automatically learn subgoal representations in an end-to-end manner along with bi-level policies~\cite{vezhnevets2017feudal,dilokthanakul2019feature,zhang2020generating}, making them more generic. However, they often overlook the temporal coherence in GCHRL. 
Temporal coherence refers to the hierarchical relationship in time between different levels of control in a system. The high-level controller operates at a slower timescale and focuses on radical changes in environmental states associated with the completion of the source task, while the low-level controller operates at a quicker timescale and focuses on modest changes in environmental states associated with completing subtasks. Approaches~\cite{li2020learning,li2021active} that consider the temporal abstraction perspective have shown great promise in improving exploration efficiency in GCHRL.

The subgoal selection strategy is another crucial component of GCHRL. The balance of exploration and exploitation and the final performance are significantly impacted. 
If the high-level controller always selects subgoals that have ever yielded high rewards, it may miss out on potentially greater rewards.
One line of work uses neural networks trained with environmental rewards as subgoal selection strategies~\cite{vezhnevets2017feudal,nachum2018data,zhang2020generating}, which enable efficient exploitation but often suffer from weak exploration.
Another line implements high-level policies as planners, achieving efficient exploitation. They build environment graphs or trees as world descriptors based on state transitions, and subgoals are generated by planning on these world descriptors~\cite{eysenbach2019search,emmons2020sparse}. 
However, this line typically builds graphs in state space, where the complexity of graph construction grows exponentially with the dimension of the state.

This paper proposes a novel method to balance exploration and exploitation in \textbf{HI}erarchical reinforcement learning via dynamically building \textbf{L}atent \textbf{L}andmark graphs (HILL). 
HILL introduces a negative-power contrastive representation learning objective to train a subgoal representation function that considers temporal coherence. 
The learned representations serve as possible subgoals, and landmarks are identified to maximize coverage of the explored latent space. Latent landmark graphs are built based on these landmarks, and two measures are defined on the graphs: a novelty measure upon nodes to encourage exploring novel landmarks and a utility measure upon edges to estimate the benefits of landmark transitions to task completion. HILL balances exploration and exploitation by simultaneously considering both measures and choosing the most valuable landmark as the subgoal.
Empirical results demonstrate that HILL outperforms state-of-the-art (SOTA) baselines in numerous challenging continuous control tasks with sparse rewards based on the MuJoCo simulator~\cite{todorov2012mujoco}. 
Furthermore, we conduct visualization analyses and ablation studies to verify the significance of the various components of HILL.
We highlight the main contributions below:
\begin{itemize}
\item We introduce a contrastive representation learning objective to train latent subgoal representations that comply with the temporal coherence in GCHRL.
\item We propose a latent landmark graph structure built on learned subgoal representations. Based on the graphs, we present a subgoal selection strategy that well tackles the exploration-exploitation dilemma in GCHRL.
\item Empirical results demonstrate the superiority of our method over SOTA baselines in numerous continuous control tasks with sparse rewards.
\end{itemize}

\section{Related Work}
\textbf{Subgoal Representation Learning.} 
Learning effective subgoal representations remains a major challenge in GCHRL.
Previous works either pre-define bottleneck states as subgoals~\cite{kulkarni2016hierarchical,nachum2018data} or use external rewards as supervision to train the high-level policy to generate subgoals~\cite{vezhnevets2017feudal,nachum2019does,levy2019learning}. The former requires task-specific knowledge and lacks generalization, while the latter results in challenging exploration and low training efficiency.
Recent works use variational autoencoders~\cite{pere2018unsupervised,nair2019hierarchical} to extract low-dimensional features from observations. However, such features may include redundant information. Instead, it is more effective to focus on the key aspects of variation that are essential for decision-making.
\citet{ghosh2018learning} learn actionable representations that emphasize state factors inducing significant differences in corresponding actions and de-emphasize irrelevant features.
Two recent studies \cite{li2020learning,li2021active} use slow feature analysis methods~\cite{wiskott2002slow} to extract useful features as subgoals. 
However, they use $L^2$ norm as a distance estimation when calculating triplet loss, which may lead to degenerate distributions~\cite{li2020focal}.
In contrast, our method uses a negative-power contrastive representation learning objective to learn informative and robust subgoal representations.

\textbf{Explore and Exploit in GCHRL.} 
Efficient exploration in GCHRL can be achieved through a variety of methods, such as scheduling between intrinsic and extrinsic task policies \cite{zhang2019scheduled}, restricting the high-level action space to reduce exploration space \cite{zhang2020generating}, designing transition-based intrinsic rewards \cite{machado2020count}, or using curiosity-driven intrinsic rewards \cite{roder2020curious}. However, these methods rarely consider exploitation efficiency after sufficient exploration.
Efficient exploitation in GCHRL is typically achieved through planning~\cite{yamamoto2018hierarchical,li2022hierarchical}, where high-level graphical planners have shown great promise~\cite{eysenbach2019search}. 
Some works use previously seen states in a replay buffer as graph nodes to generate subgoals through graph search. 
For example, \citet{shang2019learning} propose a method that unsupervisedly discovers the world graph and integrates it to accelerate GCHRL, while \citet{jin2021graph} use an attention mechanism to select subgoals based on the graphs.
However, these methods construct graphs in the state space, where the graphs can be challenging to be built in high-dimensional state spaces.
While \citet{zhang2021world} leverage an auto-encoder with reachability constraints to learn a latent space and generate latent landmarks through clustering in this space, they ultimately decode the cluster centroids as landmarks, suggesting that they still plan in the state space.
In contrast, our method builds graphs in the learned latent space, making it more generic and adaptable to state spaces of varying dimensions.

\begin{figure*}[ht]{
\centering
\includegraphics[width=0.99\textwidth]{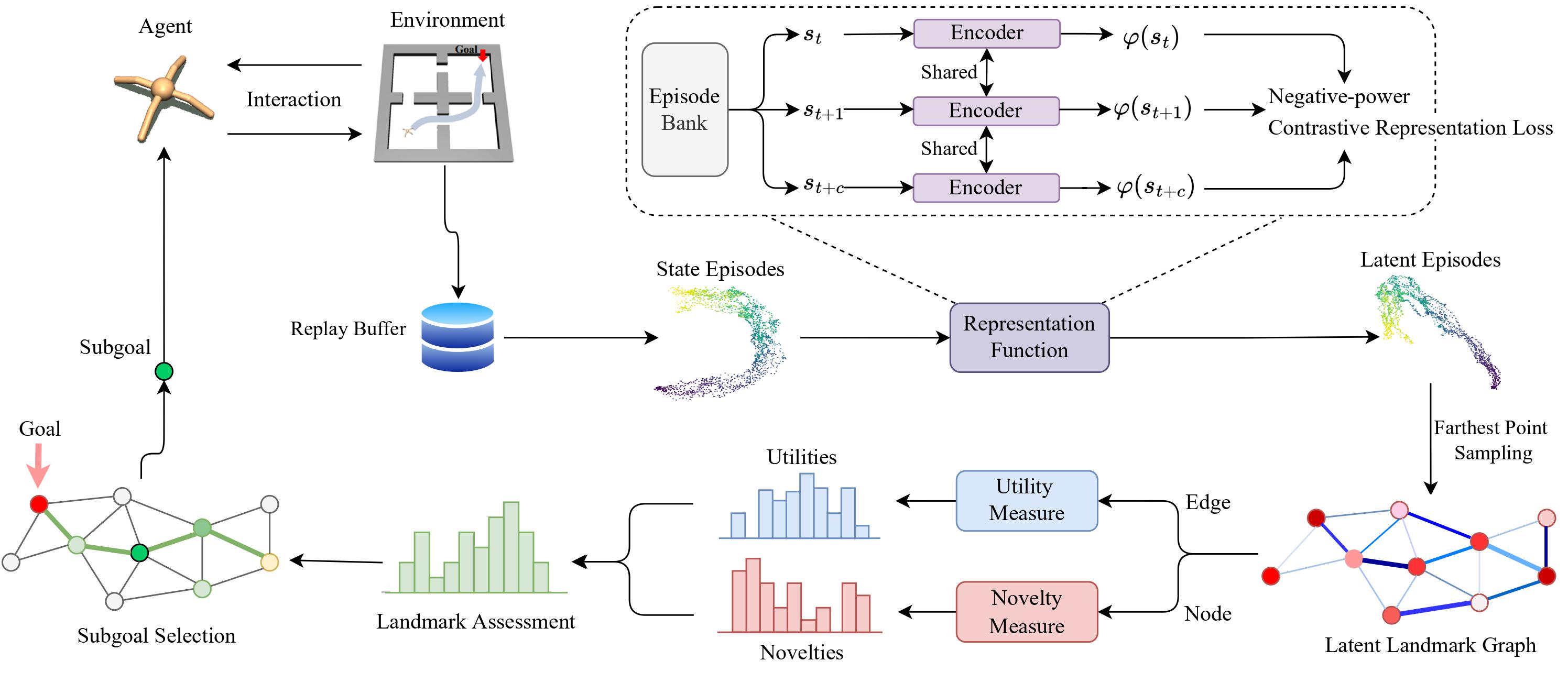}\vspace{-10pt}
\caption{An overview of HILL. HILL learns subgoal representations through a negative-power contrastive learning objective and selects subgoals by building latent landmark graphs. HILL identifies landmarks that maximize the explored latent space's coverage and constructs latent landmark graphs based on them. HILL then measures novelties of nodes to encourage exploring novel landmarks and utilities of edges to estimate the benefits of transitions between landmarks to task completion. Finally, HILL uses a balanced subgoal selection strategy to tackle the exploration-exploitation dilemma by jointly considering both measures and selecting the most valuable landmark as the next subgoal. The representation function and bi-level policies are trained jointly online.}\vspace{-10pt}
\label{fig:framework}}
\end{figure*}

\section{Preliminaries}
\subsection{Goal-Conditioned Hierarchical Reinforcement Learning} 
A finite-horizon, subgoal-augmented Markov Decision Process can be described as a tuple $\langle \mathcal{S},\mathcal{G},\mathcal{A},\mathcal{P},\mathcal{R},\gamma\rangle$, where $\mathcal{S}$ is a state space, $\mathcal{G}$ is a subgoal space, $\mathcal{A}$ is an action space, $\mathcal{P}:\mathcal{S}\times\mathcal{A}\times\mathcal{S}\rightarrow [0,1]$ is an environmental transition function, $\mathcal{R}:\mathcal{S} \times \mathcal{A} \rightarrow \mathbb{R}$ is a reward function, and $\gamma\in [0,1)$ is a discount factor. 
We consider a two-level GCHRL framework, which comprises a high-level policy $\pi_{\theta_h}(g|s)$ over subgoals given states and a low-level policy $\pi_{\theta_l}(a|s,g)$ over actions given states and subgoals.
The policy $\pi_{\theta_h}$ operates at a slower timescale and samples a subgoal $g_t \in \mathcal{G}$ when $t \equiv 0$ (mod $c$), for fixed $c$. 
The policy $\pi_{\theta_h}$ is trained to optimize the expected cumulative discounted environmental rewards $\mathbb{E}_{\pi_{\theta_h},\pi_{\theta_l}}\left[ \sum_{t=0}^{\infty}\gamma^t\mathcal{R}(s_t,a_t) \right]$. 
The policy $\pi_{\theta_l}$ selects an action $a_{t} \sim \pi_{\theta_l}(\cdot|s_t,g_t)$ at every time step and is intrinsically rewarded with $r_t^l(g_t, \varphi(s_{t}))=-D(g_t, \varphi(s_{t}))$ to reach $g_t$, where $D$ is a distance function. 
Following previous work~\cite{li2021active}, we employ $L2$ distance as $D$ to provide dense non-zero rewards, thus accelerating low-level policy learning.
The policy $\pi_{\theta_l}$ is trained to optimize the expected cumulative intrinsic rewards $\mathbb{E}_{\pi_{\theta_l}}\left[ \sum_{t=ci}^{ci+c-1}r_t^l \right],i=0,1,2,\dots$.




\subsection{Universal Value Function Approximator}
We use an off-policy algorithm Soft Actor-Critic (SAC)~\cite{haarnoja2018soft} as the base RL optimizer of both levels. The low-level critic is implemented as a Universal Value Function Approximator (UVFA)~\cite{schaul2015universal}, which extends the concept of value function approximation to include both states and subgoals as inputs. Theoretically, a sufficiently expressive UVFA can identify
the underlying structure across states and subgoals, and thus generalize to any subgoal. 

We define a pseudo-discount function $\sigma: S \rightarrow [0,1]$, which takes the double role of state-dependent discounting, and of soft termination, in the sense that $\sigma(s)=0$ if and only if $s$ is a terminal state according to the subgoal $g$. 
UVFAs can then be formalized as
$V(s, g)\approx  V_{g, \pi_{\theta_l}^*}(s)$, where $\pi_{\theta_l}^*$ is the optimal low-level policy and $V_{g,\pi_{\theta_l}}$ is a general value function representing the expected cumulative pseudo-discounted future pseudo-return for each $g$ and any policy $\pi_{\theta_l}$:
\begin{equation}
V_{g,\pi_{\theta_l}}(s)=\mathbb{E}\bigg[\sum_{t=0}^{\infty}r_t^l(g, \varphi(s_{t}))\prod_{k=0}^{t}\sigma(s_k)\bigg|s_0=s\bigg].
\end{equation}
$V(s, g)$ is usually implemented by a neural network for better generalization and is trained by the Bellman equation in a bootstrapping way. 

\section{Method}
We introduce HILL in detail in this section, and its main structure is depicted in Figure \ref{fig:framework}.
\subsection{Contrastive Subgoal Representation Learning}
We define a subgoal representation function $\varphi: \mathcal{S} \rightarrow \mathbb{R}^d$ which abstracts states $s \in \mathcal{S}$ to latent representations $z \in \mathcal{Z}$.
Assuming the high-level controller selects a subgoal every $c$ time steps. According to the temporal coherence in GCHRL, the representations of high-level adjacent states (e.g., $\varphi(s_{t})$ and $\varphi(s_{t+c})$) are distinguishable, while those of low-level adjacent states (e.g., $\varphi(s_{t})$ and $\varphi(s_{t+1})$) are relatively similar. To train $\varphi$, we define a negative-power contrastive representation learning objective as follows:
\begin{equation}
\begin{aligned}
L_{c}(s_t, s_{t+1}, s_{t+c}) = &||\varphi(s_{t}) - \varphi(s_{t+1})||_2^2\\
&+ \beta\cdot\frac{1}{||\varphi(s_{t})-\varphi(s_{t+c})||_2^n+\epsilon}
 \label{equ:contrastiveloss},
\end{aligned}
\end{equation}
where $n \geq 1$, $\beta$ is a scaling factor, and $\epsilon>0$ is a small constant to avoid a zero denominator. The function $\varphi$ abstracts the high-dimensional state space into a low-dimensional information-intensive latent space (i.e., subgoal space), significantly reducing the exploration difficulty.

The function $\varphi$ is trained with the bi-level policies and value functions jointly. To alleviate the non-stationarity caused by dynamically updated representations $z$, we adopt a regularization function $L_r$ similar to HESS \cite{li2021active} to restrict the change of representations that already well-fit Equation~\ref{equ:contrastiveloss}:
\begin{equation}
\begin{aligned}
\label{reg}
L_r(s_t) = ||\varphi(s_t) - \varphi_{old}(s_t)||_2.
\end{aligned}
\end{equation}
We maintain a replay buffer $\mathcal{B}_p$ to record the newest losses of the sampled triplets calculated by Equation~\ref{equ:contrastiveloss}.  
When updating $\varphi$, we sample the top $k$ triplets with the smallest losses and calculate their $L_r$ to regularize its update. The overall subgoal representation learning loss is:
\begin{equation}
\begin{aligned}
L_\varphi(s) = \mathbb{E}_{s_t \sim \mathcal B_l}[L_{c}(s_t, s_{t+1}, s_{t+c})] + \mathbb{E}_{s_t^{\prime} \sim \mathcal B_p }[L_{r}(s_t^{\prime})],
\label{equ:neg}
\end{aligned}
\end{equation}
where $\mathcal{B}_l$ is a replay buffer containing historical episodes.

\subsection{Building Latent Landmark Graphs}
At time steps $t \equiv 0$ (mod $c$), we build a latent landmark graph by randomly sampling $K$ states from $\mathcal{B}_l$, abstracting them into representations using $\varphi$, and store the state and its representation in pairs into a temporary buffer $\mathcal{B}_t$. 
We then build a latent landmark graph that balances subgoal selection for effective exploration-exploitation trade-offs.
Every time a new graph is built, $\mathcal{B}_t$ is reset.
 
\subsubsection{Novelty-guided Exploration Based on Nodes}
\textbf{Sampling Nodes.}
We use the Farthest Point Sampling (FPS)~\cite{vassilvitskii2006k} method to select a collection of representations from $\mathcal{B}_t$ that maximize the coverage of the latent space explored. 
These representations, called landmarks and denoted as $\mathcal{L}_t$, serve as nodes of the latent landmark graph and optional subgoals. We also add the representation of the task goal that the agent receives at the start of the current episode, as well as the representation of the current state, to $\mathcal{L}_t$. A single landmark is denoted as $l \in \mathcal{L}_t$.

\textbf{Novelty Measure upon Nodes.}
To identify novel landmarks that can guide the agent to states it has rarely visited before, we utilize the count-based method~\cite{tang2017exploration} and define a novelty measure $N$ for any representation (including $l \in \mathcal{L}_t$). This measure estimates the expected discounted future occupancies of representations starting from a given state $s_i$:
\begin{equation}
    N(z_i) = \sum_{\mathcal{T} \in \mathcal{B}_l}\sum_{j=0}^{\lfloor (T-i)/c \rfloor}\gamma^{j} \eta(z_{i+jc}),
\label{equ:novelty}
\end{equation}
where $z_i=\varphi(s_i)$, $T$ is the length of episode $\mathcal{T}$ and $\eta$ is a hash table~\cite{charikar2002similarity} that records the historical cumulative visit counts of representations. 
By leveraging the visit history of past episodes in $\mathcal{B}_l$, $N$ realizes a long-term estimate of novelty. A landmark $l$ with smaller $N(l)$ is considered more worth exploring.
We update the hash table $\eta$ at the end of each episode.

 
\subsubsection{Utility-guided Exploitation Based on Edges}
\textbf{Connecting Nodes into Edges.}
We create two edges directed in reverse orders between each pair of latent landmarks $l_i,l_j \in \mathcal{L}_t$. Specifically, an edge $w_{i,j}$ is defined from $l_i$ to $l_j$.

\textbf{Utility Measure upon Edges.}
We define a \textit{utility} measure $U$ for any $w_{i,j}$, which estimates the benefits of transitioning from $l_i$ to $l_j$ in completing the source task:
\begin{equation}
\begin{aligned}
\label{equ:utility}
U(w_{i,j}) = \mathbb{E}_{s_x \in \mathcal{B}_l}[I(\varphi(s_x) = l_i)V(s_x,l_j)],
\end{aligned}
\end{equation}
where $I$ is an indicator function and $V$ is the low-level UVFA. 
The measure $U$ depends on the accuracy of $V$.
However, learning a UVFA that accurately estimates $V$ poses special challenges. The agent encounters limited combinations of states and subgoals $(s, g)$, which can lead to unreliable estimates for unseen combinations.
To address this issue, we consider that, starting from a point (i.e., a representation) in the latent space, representations within the neighborhood of this point are more likely to be visited, and thus, $V$ provides accurate estimates. Therefore, we obtain $U$ for non-adjacent landmarks by applying the Bellman-Ford \cite{mcquillan1977arpa} method:
\begin{equation}
\begin{aligned}
\label{equ:bellman-ford}
U(w_{i,f}) &= \mathop{\max}_{l_j}[U(w_{i,j})+U(w_{j,f})] \\
&= \mathop{\max}_{l_j,\dots,l_v}[U(w_{i,j})+\sum_{x=j}^{v-1}U(w_{x,x+1})+U(w_{v,f})],
\end{aligned}
\end{equation}
where $w_{i,f}$ and $w_{j,f}$ are the edges of non-adjacent landmarks $(l_i,l_f)$ and $(l_j,l_f)$, respectively, and $w_{i,j}$, $w_{x,x+1}$ and $w_{v,f}$ are the edges of adjacent landmarks $(l_i,l_j)$, $(l_x,l_{x+1})$ and $(l_v,l_f)$, respectively.
A landmark $l_j$ with greater $U(w_{i,j})$ is more worth exploiting. Moreover, we accelerate the convergence of $V$ using Hindsight Experience Replay (HER) \cite{andrychowicz2017hindsight}. HER smartly generates more feedback for the agent by replacing unachievable subgoals with achieved ones in the near future, thus accelerating the training process. 

\subsection{Balance Exploration and Exploitation}
We tackle the exploration-exploitation dilemma by selecting a landmark from $\mathcal{L}_t$ that well-balances $N$ and $U$ as the subgoal.
When mapped to the $[0,1]$ distribution, $U(w_{i,j})$ can be regarded as an incremental probability of success as it approximates the expected benefits of transitioning from $l_i$ to $l_j$. The distribution is as follows:
\begin{equation}
\begin{aligned}
\mathcal{P}(U(w_{i,j})) =\frac{e^{U(w_{i,j})}}{\sum_{y=1}^{Y}e^{U(w_{i,y})}},
\end{aligned}
\end{equation}
where $Y$ is the number of landmarks in $\mathcal{L}_t$. Then we define a balanced strategy based on $N$ and $\mathcal{P}$ and select a landmark with a smaller historical visit count and a larger transitional benefit as the subgoal $g_t$ of the next time interval:
\begin{equation}
\begin{aligned}
g_t = \mathop{\arg\min}\limits_{l_j\in \mathcal{L}_t} [(1-\mathcal{P}(U(w_{i^{\prime},j}))) \times N(l_j)],
\label{equ:sgseletion}
\end{aligned}
\end{equation}
where $l_{i^{\prime}} =\varphi(s_t)$ and $s_t$ is the state of the current time step.

Algorithm~\ref{alg:HILL} shows the pseudo-code of HILL.
Note that we use the subgoal selection strategy in Equation \ref{equ:sgseletion} as a high-level teacher policy and use $\pi_{\theta_h}$ modeled by a neural network as a high-level student policy.
Both policies map states to subgoals and are employed with a probability of $p$ and $1-p$, respectively. The role of the student policy is mainly to help increase randomness and prevent falling into local optimum. $p$ is initialized to 0.5 and gradually increases to 1 in the training process. 
$\theta_h$ is trained with SAC using episodes in $\mathcal{B}_l$.

\begin{algorithm}[t]
\caption{HILL algorithm}
\label{alg:HILL}
\textbf{Initialize}: $\varphi$, $\pi_{\theta_h}$, $\pi_{\theta_l}$ and $p$.
\begin{algorithmic}[1] 
\FOR {$i=1...num\_episodes$}
    \FOR {$t=0...T-1$}
        \IF {$t \equiv 0$ (mod $c$)}
            \STATE Generate a random float value $q \in [0.0, 1.0]$.
            \IF {$2q-1 \le p$ }
                \STATE Build a latent landmark graph.
                \STATE Select $g_t$ with the strategy in Equation \ref{equ:sgseletion}.
            \ELSE
                \STATE Sample $g_t$ using $\pi_{\theta_h}$.
            \ENDIF
            \STATE Update $\theta_h$ using SAC.
        \ENDIF
        \STATE $r_t, s_{t+1} \leftarrow$ execute $a_t \sim \pi_{\theta_l}(\cdot|s_t, g_t)$. 
    \ENDFOR
    \IF{$i \equiv 0$ (mod $100$)}
        \STATE{Update $\varphi$ using Equation \ref{equ:neg}.}
        \STATE Update $p$ to the newest average train success rate.
    \ENDIF
    \STATE Update $\theta_l$ and $\varphi$ using SAC.
\ENDFOR
\STATE \textbf{return} $\varphi$, $\pi_{\theta_h}$ and $\pi_{\theta_l}$.
\end{algorithmic}
\end{algorithm}

\begin{figure*}[ht]
    \centering
    \setcounter{subfigure}{0}
	\centering
	\subfigure[Ant Maze]{
		\begin{minipage}[t]{0.3\textwidth}
			\centering
			\includegraphics[width=\textwidth]{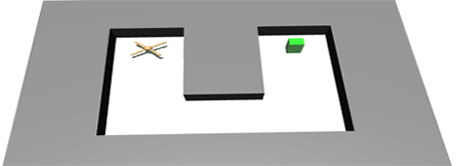}
		\end{minipage}
	}\vspace{-3pt}
	\subfigure[Ant FourRooms]{
		\begin{minipage}[t]{0.3\textwidth}
			\centering
			\includegraphics[width=\textwidth]{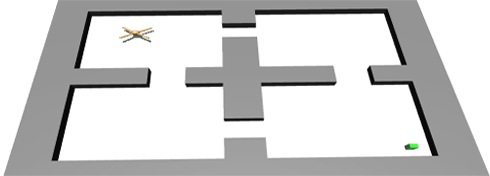}
		\end{minipage}
	}\vspace{-3pt}
	\subfigure[Ant Push]{
		\begin{minipage}[t]{0.3\textwidth}
			\centering
			\includegraphics[width=\textwidth]{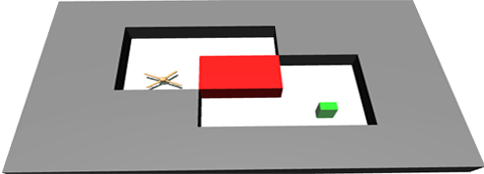}
		\end{minipage}
	}\vspace{-3pt}
	
	\subfigure[HalfCheetah Hurdle]{
		\begin{minipage}[t]{0.3\textwidth}
			\centering
			\includegraphics[width=\textwidth]{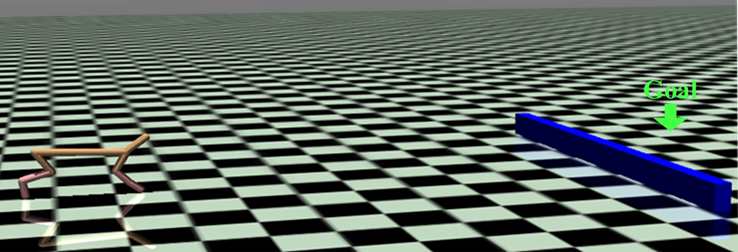}
		\end{minipage}
	}\vspace{-2pt}
	\subfigure[HalfCheetah Climbing]{
		\begin{minipage}[t]{0.3\textwidth}
			\centering
			\includegraphics[width=\textwidth]{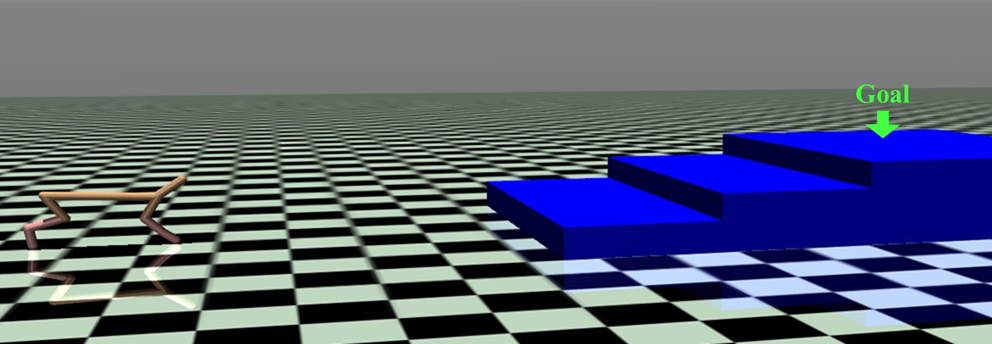}
		\end{minipage}
	}\vspace{-2pt}
	\subfigure[HalfCheetah Ascending]{
		\begin{minipage}[t]{0.3\textwidth}
			\centering
			\includegraphics[width=\textwidth]{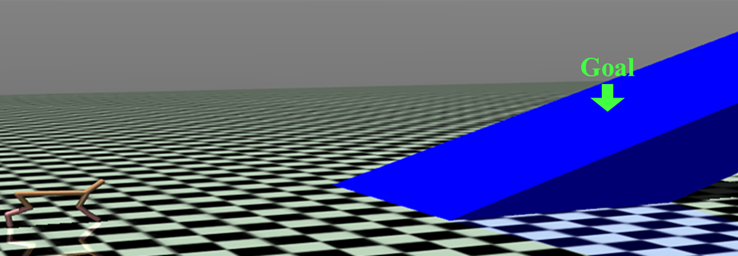}
		\end{minipage}
	}\vspace{-2pt}
	\centering
	\caption{The continuous control environments with sparse rewards for evaluating HILL and baselines.}\vspace{-5pt}
	\label{fig:env}
\end{figure*}

\begin{figure*}[ht]
    \centering
    \subfigure{
		\begin{minipage}[t]{0.99\textwidth}
			\centering
			\includegraphics[width=\textwidth]{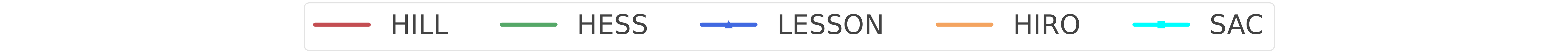}
		\end{minipage}
	}\vspace{-5pt}%

    \setcounter{subfigure}{0}
	\centering
	\subfigure[Ant Maze]{
		\begin{minipage}[t]{0.3\textwidth}
			\centering
			\includegraphics[width=\textwidth]{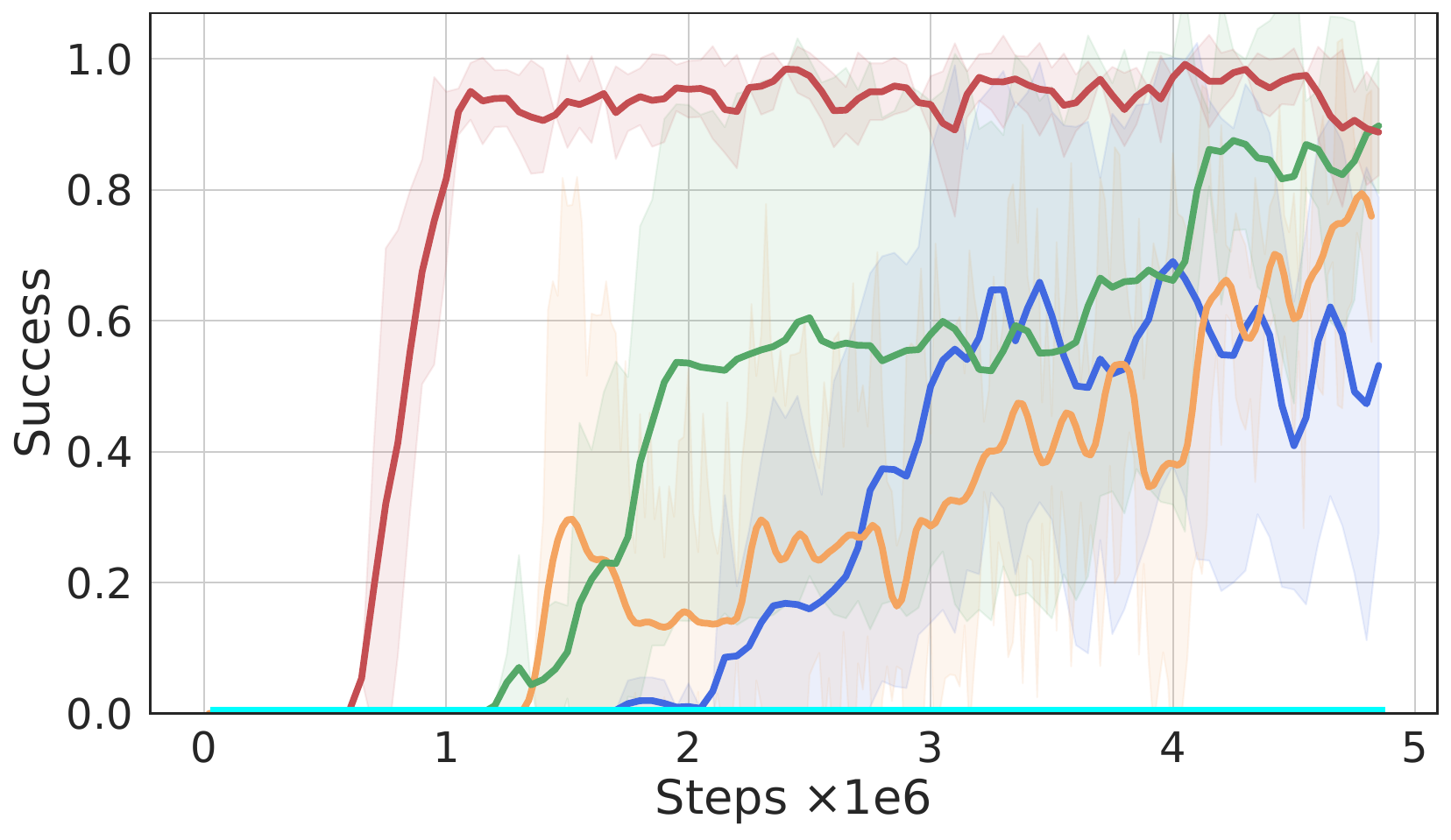}
		\end{minipage}
	}\vspace{-2pt}
	\subfigure[Ant FourRooms]{
		\begin{minipage}[t]{0.3\textwidth}
			\centering
			\includegraphics[width=\textwidth]{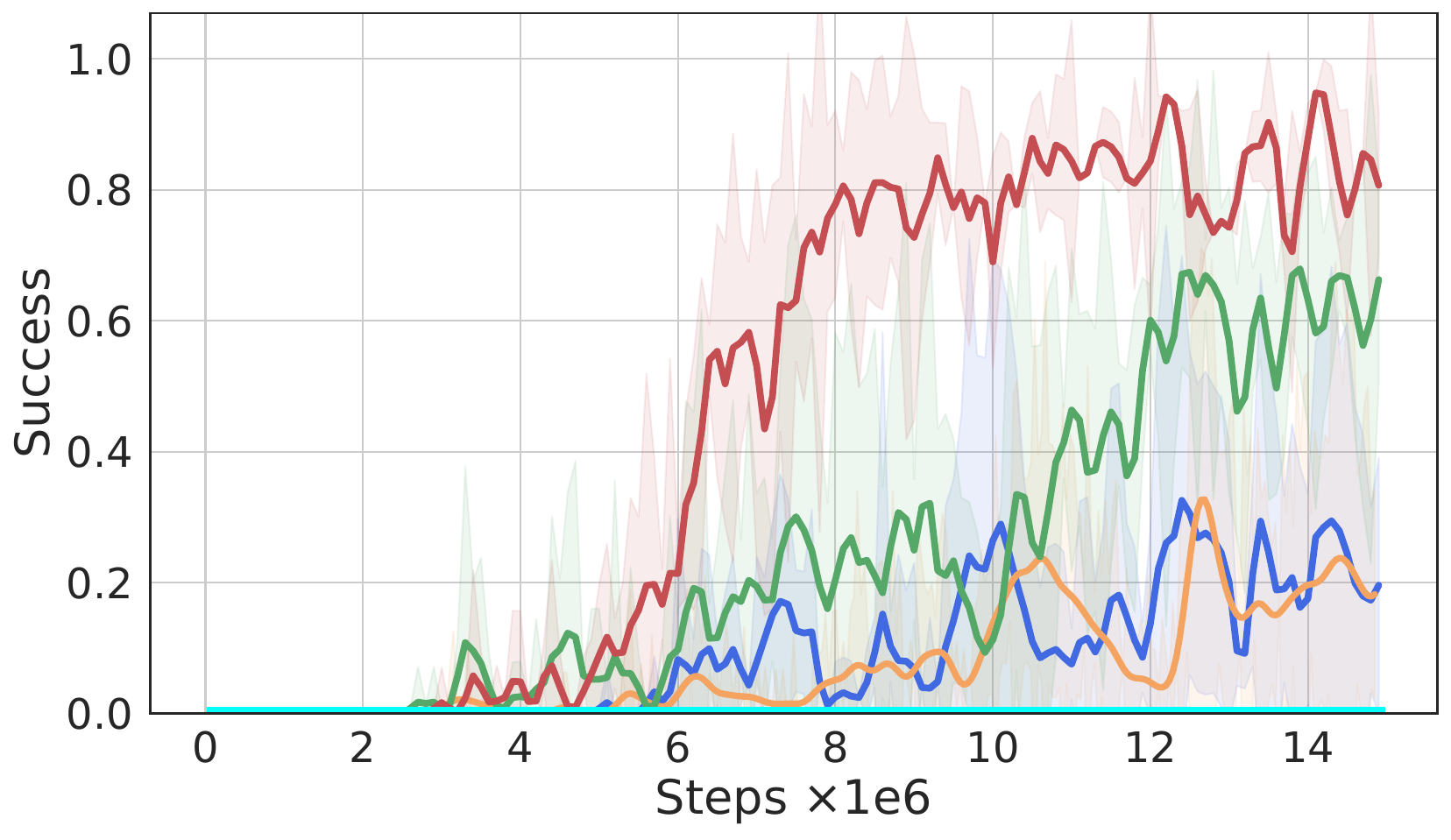}
		\end{minipage}
	}\vspace{-2pt}
	\subfigure[Ant Push]{
		\begin{minipage}[t]{0.3\textwidth}
			\centering
			\includegraphics[width=\textwidth]{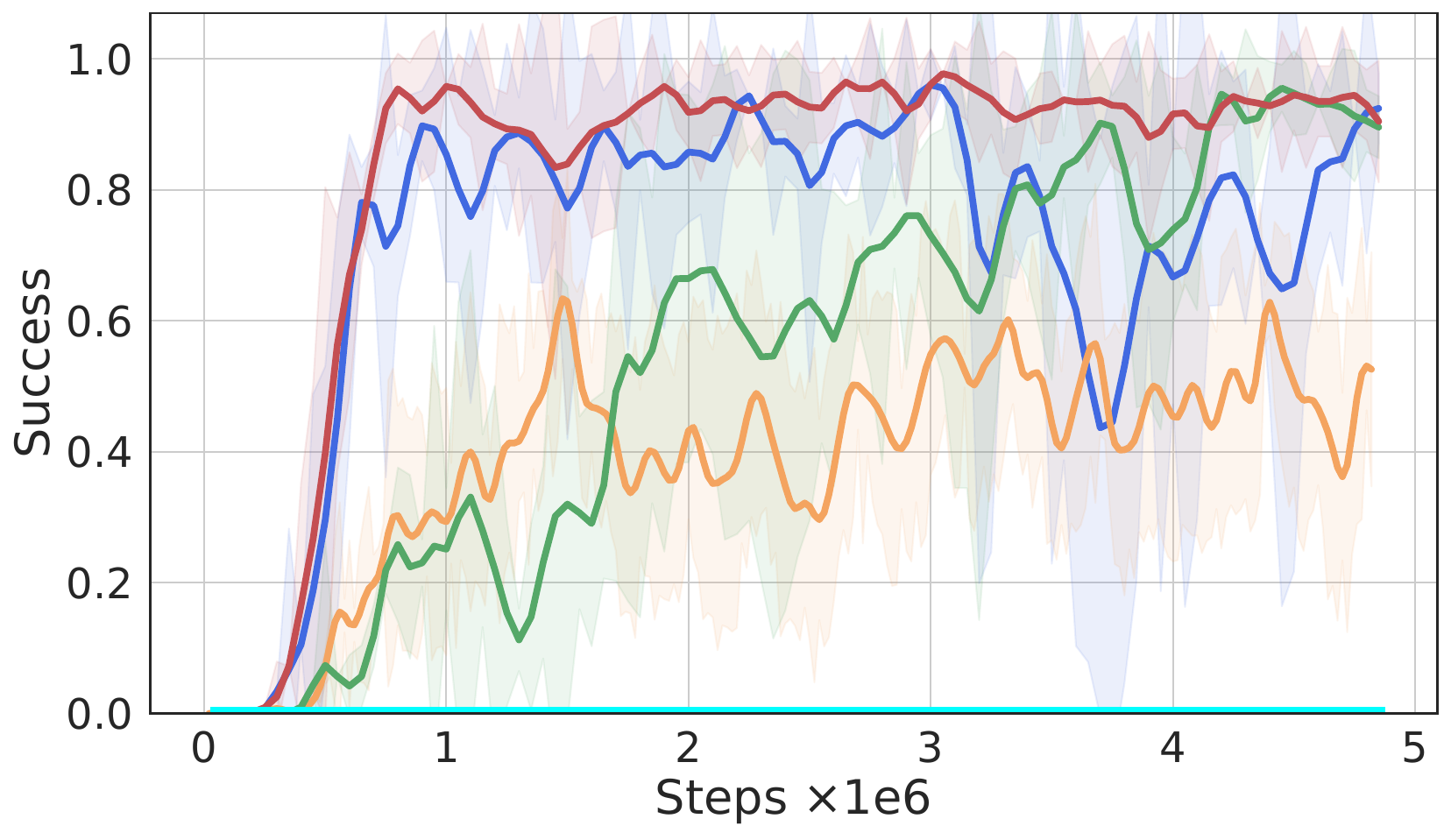}
		\end{minipage}
	}\vspace{-2pt}
	
	\subfigure[HalfCheetah Hurdle]{
		\begin{minipage}[t]{0.3\textwidth}
			\centering
			\includegraphics[width=\textwidth]{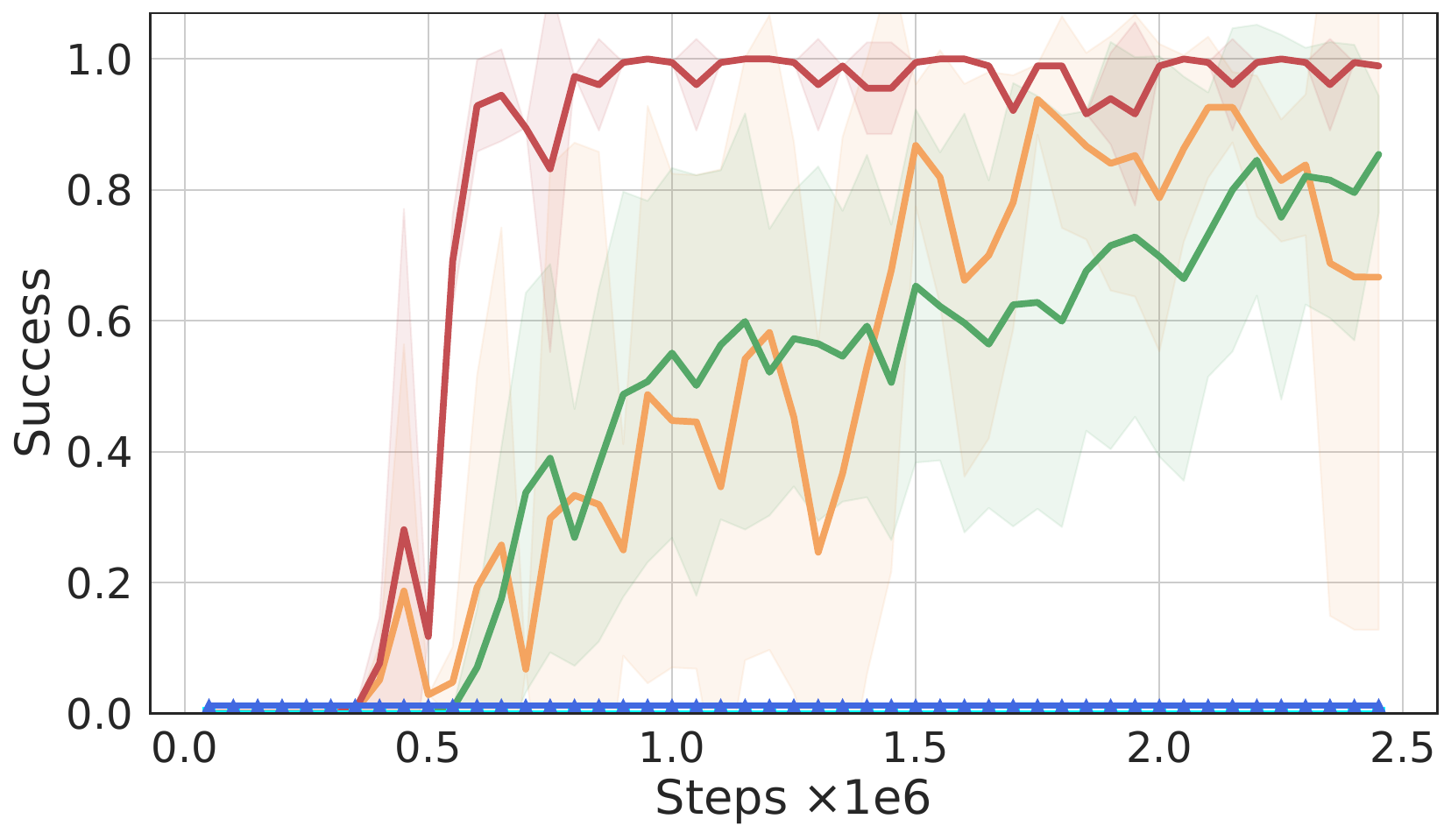}
		\end{minipage}
	}\vspace{-2pt}
	\subfigure[HalfCheetah Climbing]{
		\begin{minipage}[t]{0.3\textwidth}
			\centering
			\includegraphics[width=\textwidth]{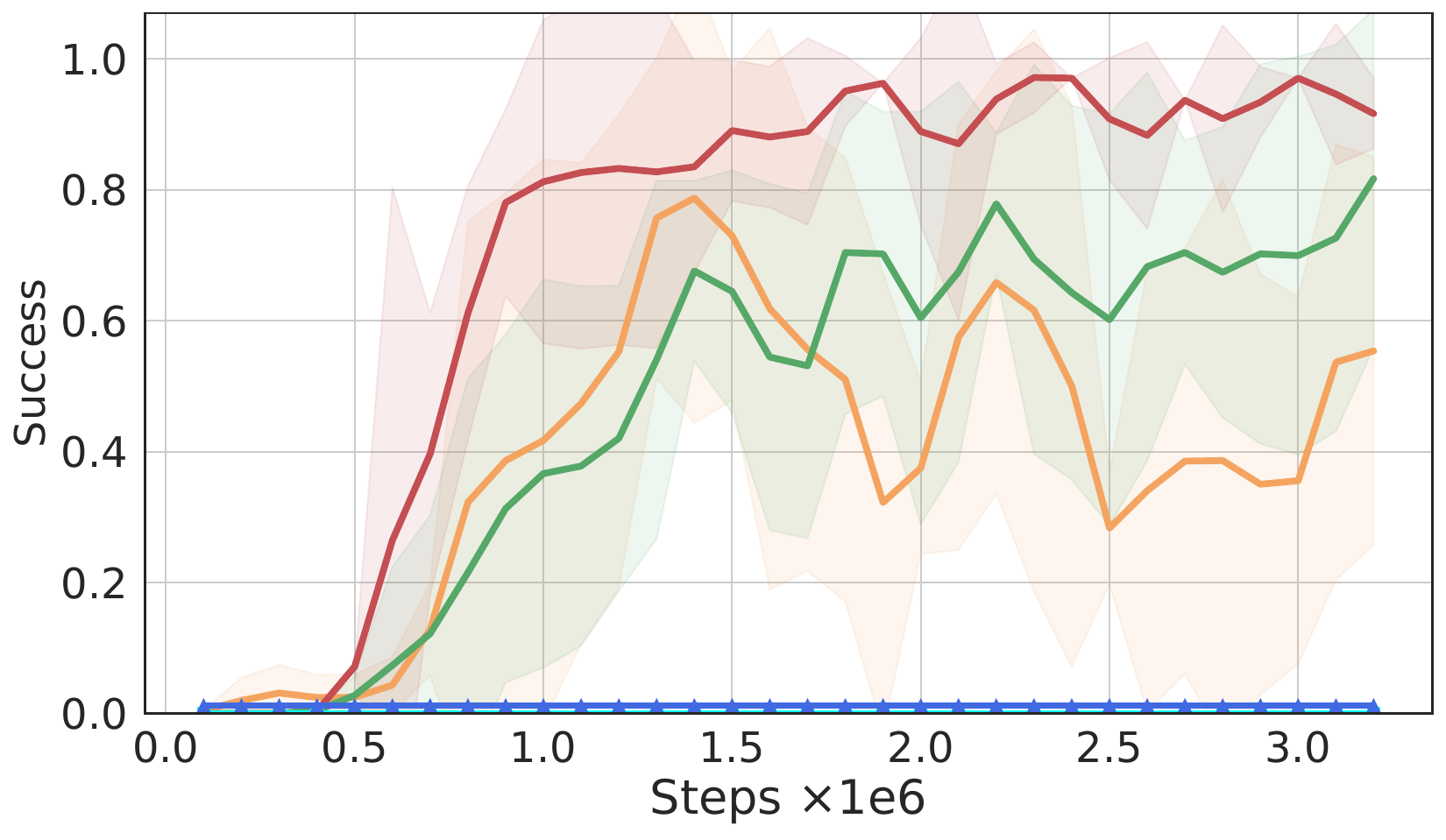}
		\end{minipage}
	}\vspace{-2pt}
	\subfigure[HalfCheetah Ascending]{
		\begin{minipage}[t]{0.3\textwidth}
			\centering
			\includegraphics[width=\textwidth]{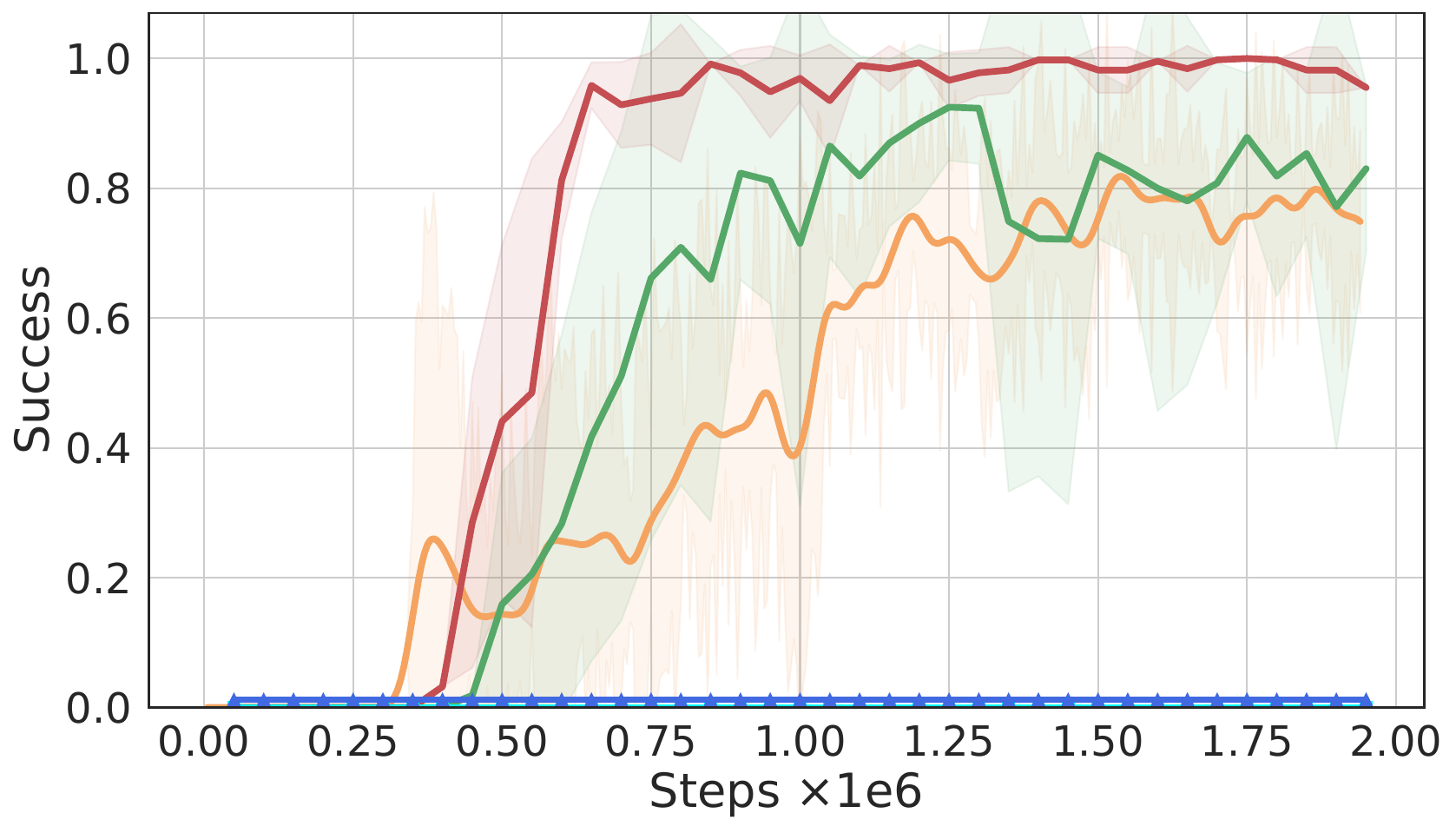}
		\end{minipage}
	}\vspace{-2pt}
	\centering
	\caption{Learning curves of HILL and baselines on MuJoCo tasks. The $x$-axis shows the time step of training, and the y-axis shows the average test success rate of $10$ episodes. Each line is the mean of $5$ runs with shaded regions corresponding to the $95\%$ confidence interval.}
	\label{main results}\vspace{-7pt}
\end{figure*}

\begin{figure*}[ht]{
\centering
\includegraphics[width=0.99\textwidth]{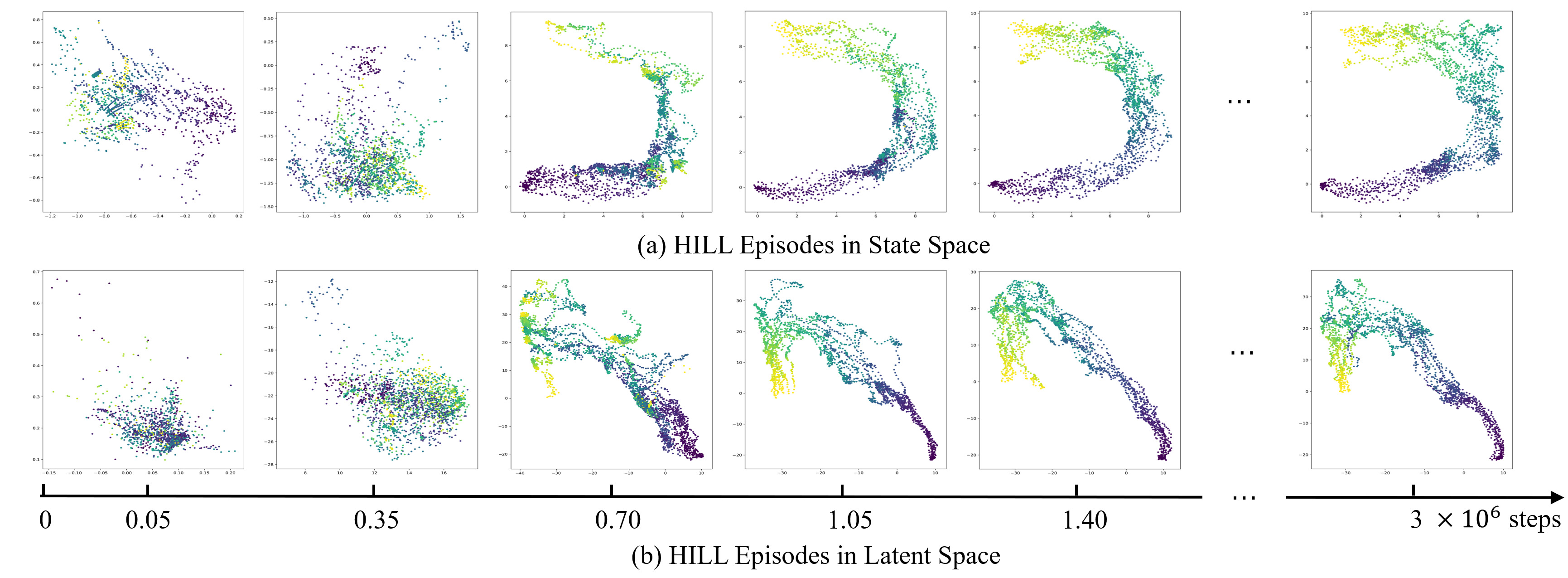}\vspace{-10pt}
\caption{Episodes in the state space and representations in the latent space at different learning stages in the Ant Maze task.}\vspace{-8pt}
\label{fig:representation}}
\end{figure*}

\section{Experiments}
\begin{figure}[ht]
	\centering
	\subfigure[State DIST. ($0.5M$ steps)]{
		\begin{minipage}[t]{0.48\linewidth}
			\centering
			\includegraphics[width=\linewidth]{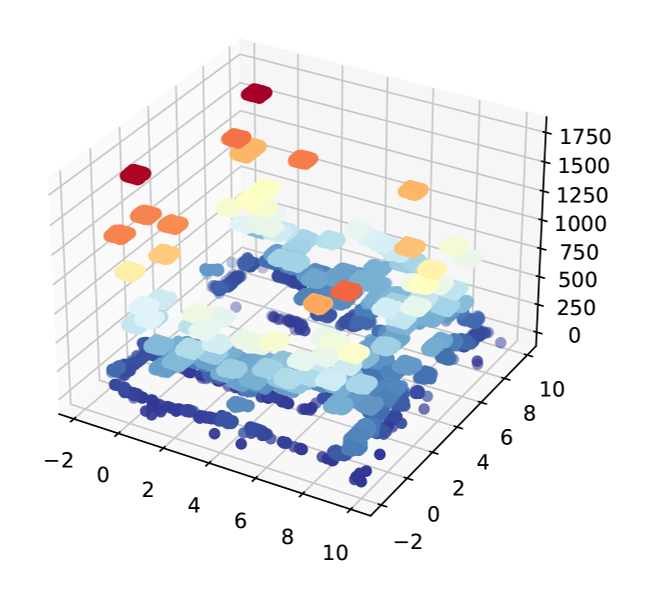}
		\end{minipage}
	}\vspace{-5pt}%
    \subfigure[State DIST. ($1.5M$ steps)]{
		\begin{minipage}[t]{0.48\linewidth}
			\centering
			\includegraphics[width=\linewidth]{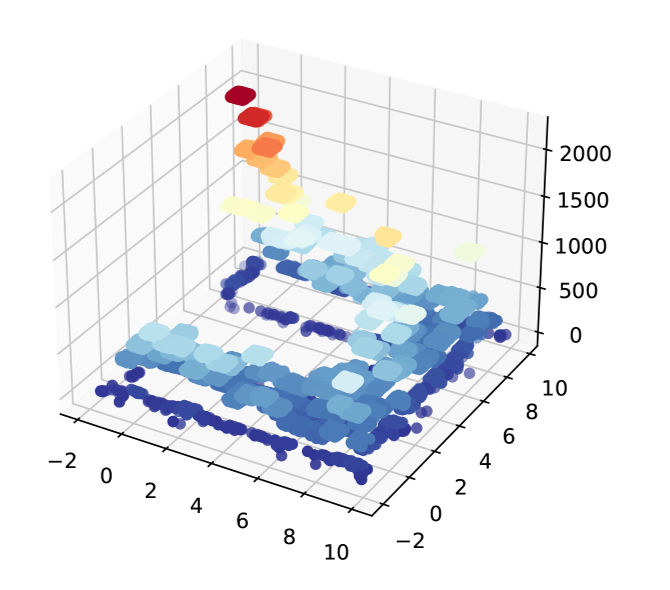}
		\end{minipage}
	}\vspace{-5pt}%
 
	\subfigure[REP. DIST. ($0.5M$ steps)]{
		\begin{minipage}[t]{0.48\linewidth}
			\centering
			\includegraphics[width=\linewidth]{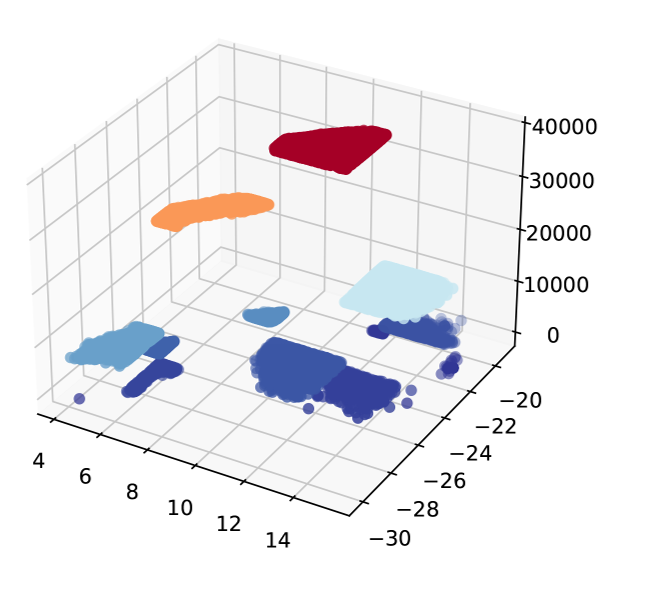}
		\end{minipage}
	}
	\subfigure[REP. DIST. ($1.5M$ steps)]{
		\begin{minipage}[t]{0.48\linewidth}
			\centering
			\includegraphics[width=\linewidth]{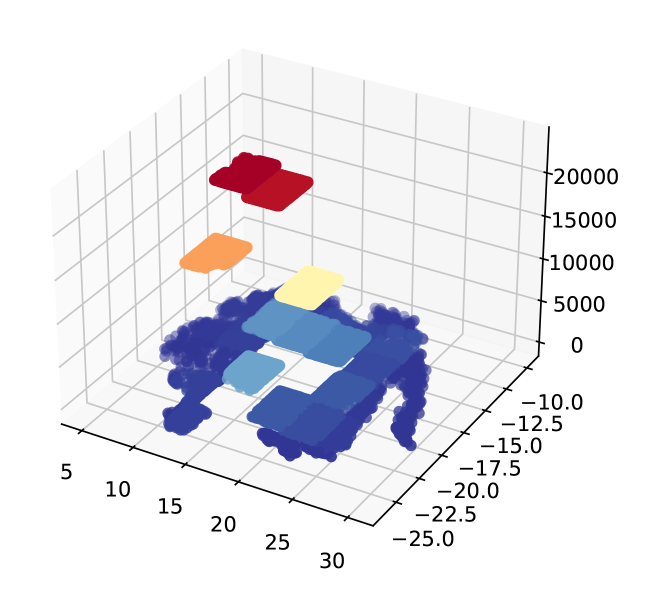}
		\end{minipage}
	}
	\centering
	\caption{Visualization of exploration and exploitation in the Ant Maze task at 0.5 and 1.5 million steps, respectively. The $x,y$-axes are positions in the corresponding space, and the $z$-axis is the visit counts statistically obtained from the replay buffer with a capacity of $10^5$. ``REP.'' denotes the word ``representation'' and ``DIST.'' denotes the word ``distribution''.}\vspace{-12pt}
	\label{fig:exploration}
\end{figure}

\subsection{Environmental Settings}
We evaluate HILL on a suite of challenging continuous control tasks based on the MuJoCo simulator~\cite{todorov2012mujoco}. These tasks require the agent to master a combination of locomotion and object manipulation skills.
The reward is sparse: the agent gets $0$ when reaching the goal and $-1$ otherwise. During training, the agent is initialized with random positions and is tested under the most challenging setting to reach the other side of the environment. 
We conduct experiments on six environments in MuJoCo, and their visualizations are presented in Figure~\ref{fig:env}.
The maximum episode length is limited to 500 steps for the Ant Maze, Ant Push, HalfCheetah Hurdle, and HalfCheetah Ascending tasks and 1000 steps for the Ant FourRooms and HalfCheetah Climbing tasks.

\begin{figure}[t]
	\centering
	\subfigure{
		\begin{minipage}[t]{0.98\linewidth}
			\centering
			\includegraphics[width=\linewidth]{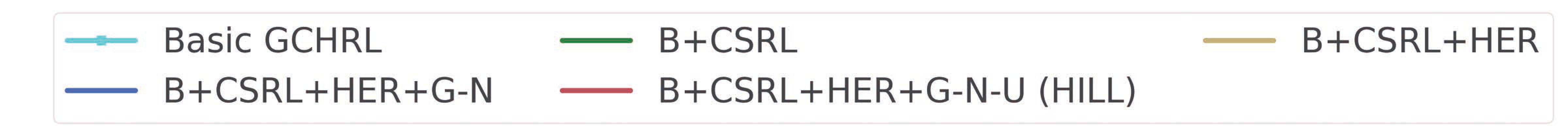}
		\end{minipage}
	}\vspace{-10pt}

    \setcounter{subfigure}{0}
	\centering
	\subfigure[Ant Maze]{
		\begin{minipage}[t]{0.46\linewidth}
			\centering
			\includegraphics[width=\linewidth]{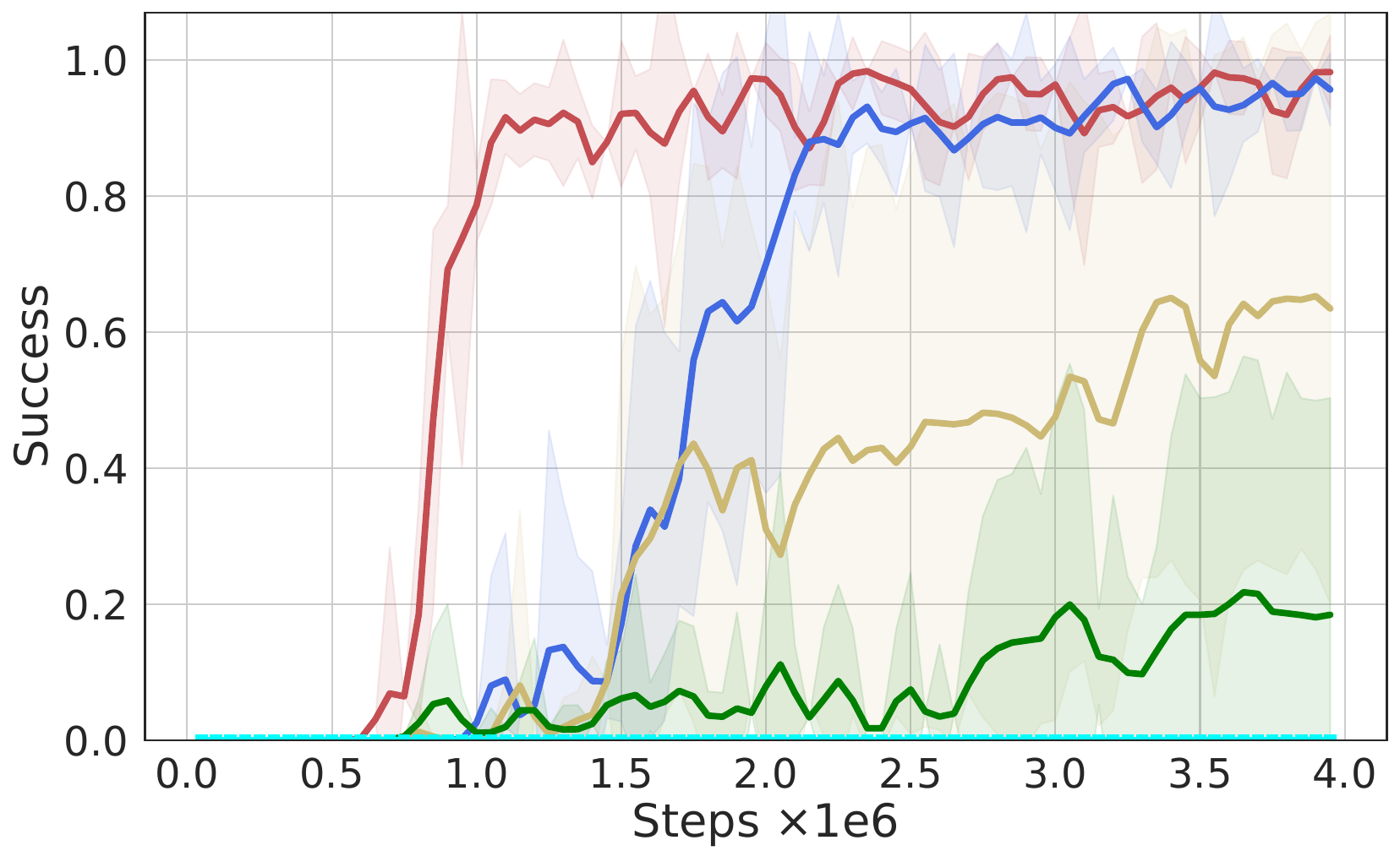}
		\end{minipage}
	}%
	\subfigure[Ant Push]{
		\begin{minipage}[t]{0.46\linewidth}
			\centering
			\includegraphics[width=\linewidth]{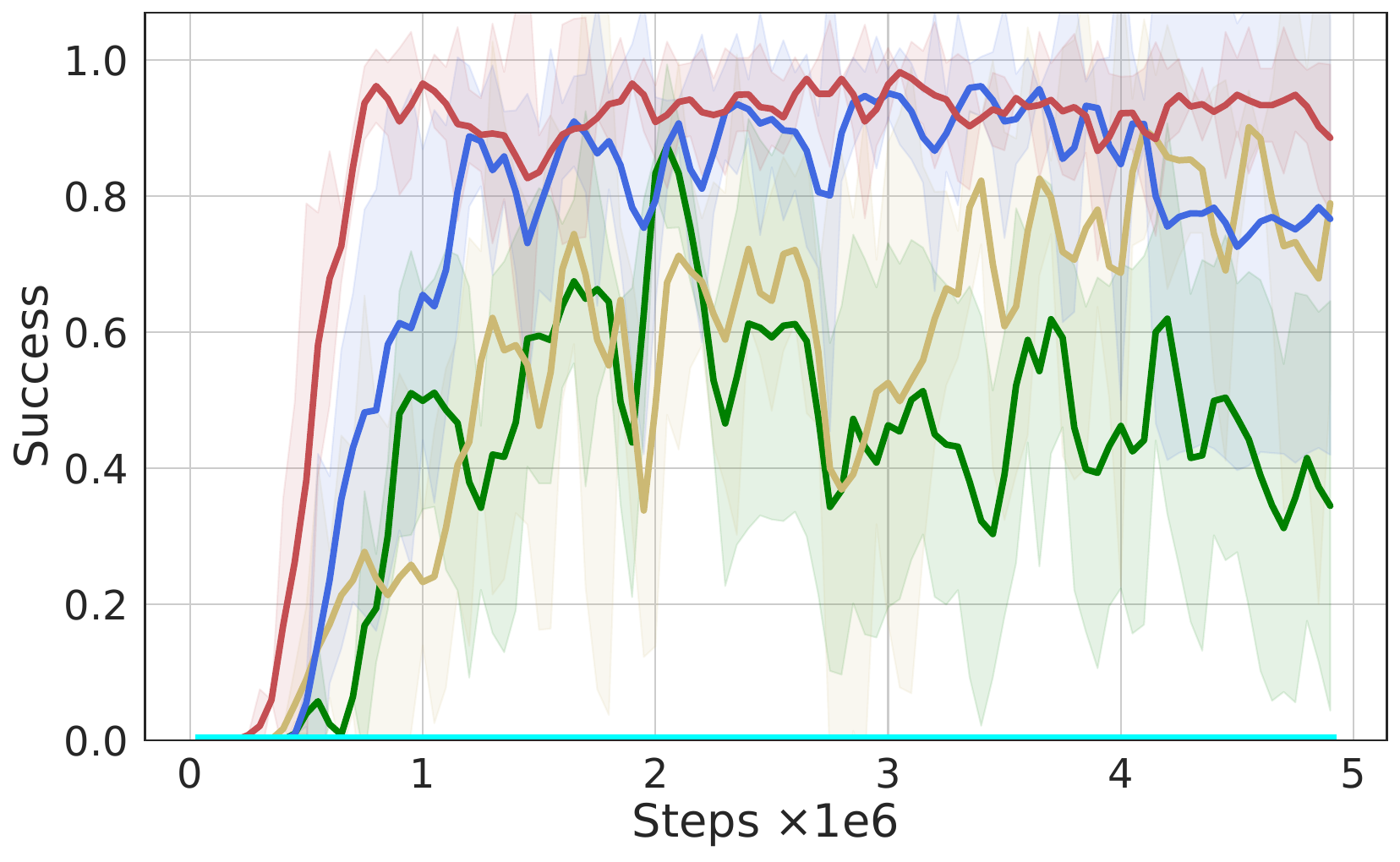}
		\end{minipage}
	}
	\centering	\caption{Ablation studies of critical modules. ``B'' denotes ``Basic GCHRL'', ``CSRL'' denotes ``Contrastive Subgoal Representation Learning'', ``G-N'' denotes ``building Graphs with the Novelty measure'', and ``G-N-U'' denotes ``building Graphs with the Novelty measure and the Utility measure''.}\vspace{-10pt}
	\label{fig:ablation1}
\end{figure}

\begin{figure}[ht]
	\centering
  \subfigure[Stable Ratio $k$]{
		\begin{minipage}[t]{0.47\linewidth}
			\centering
			\includegraphics[width=\linewidth]{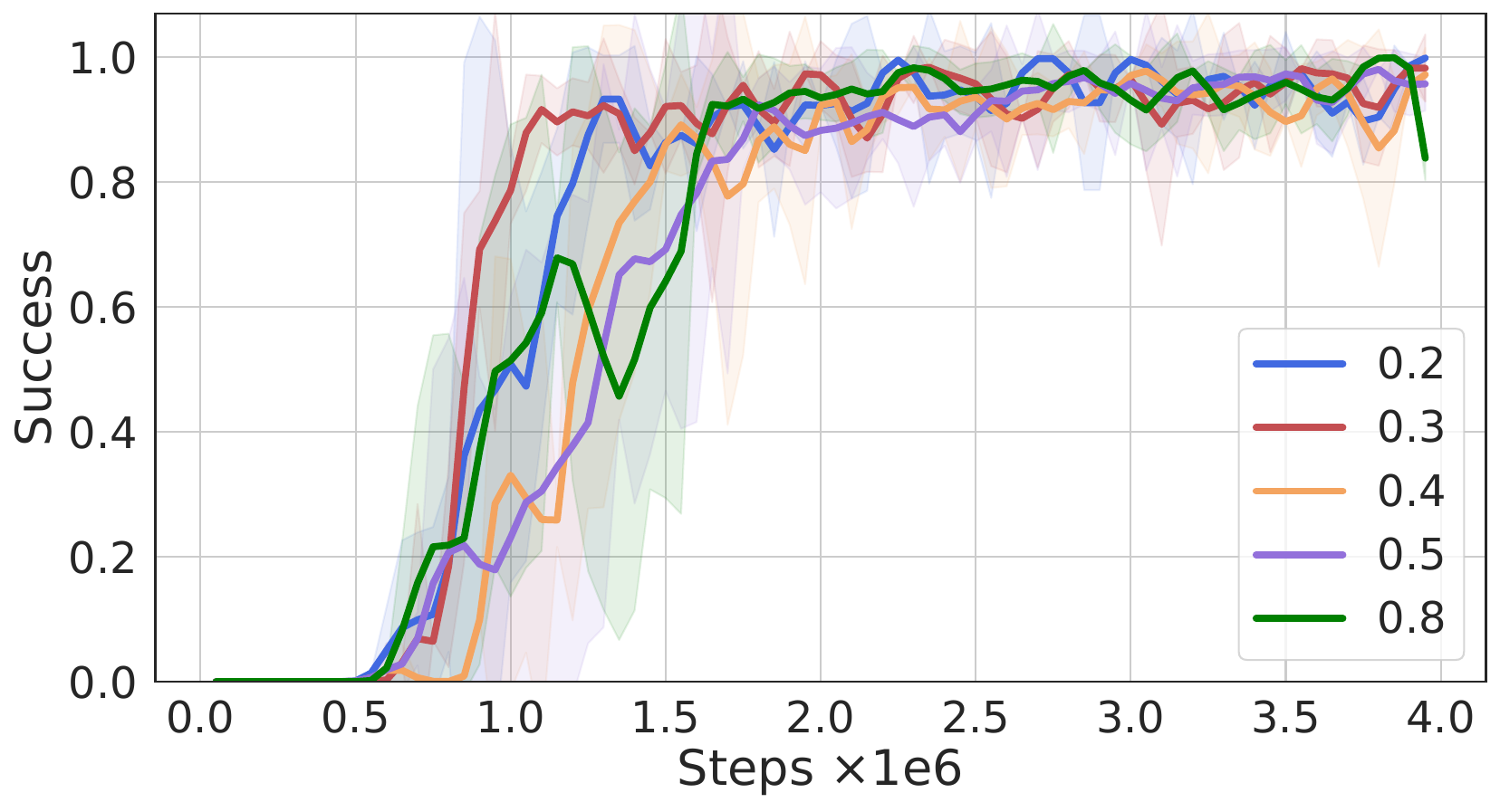}
		\end{minipage}
	}%
    \subfigure[Representation Dimension $d$]{
		\begin{minipage}[t]{0.47\linewidth}
			\centering
			\includegraphics[width=\linewidth]{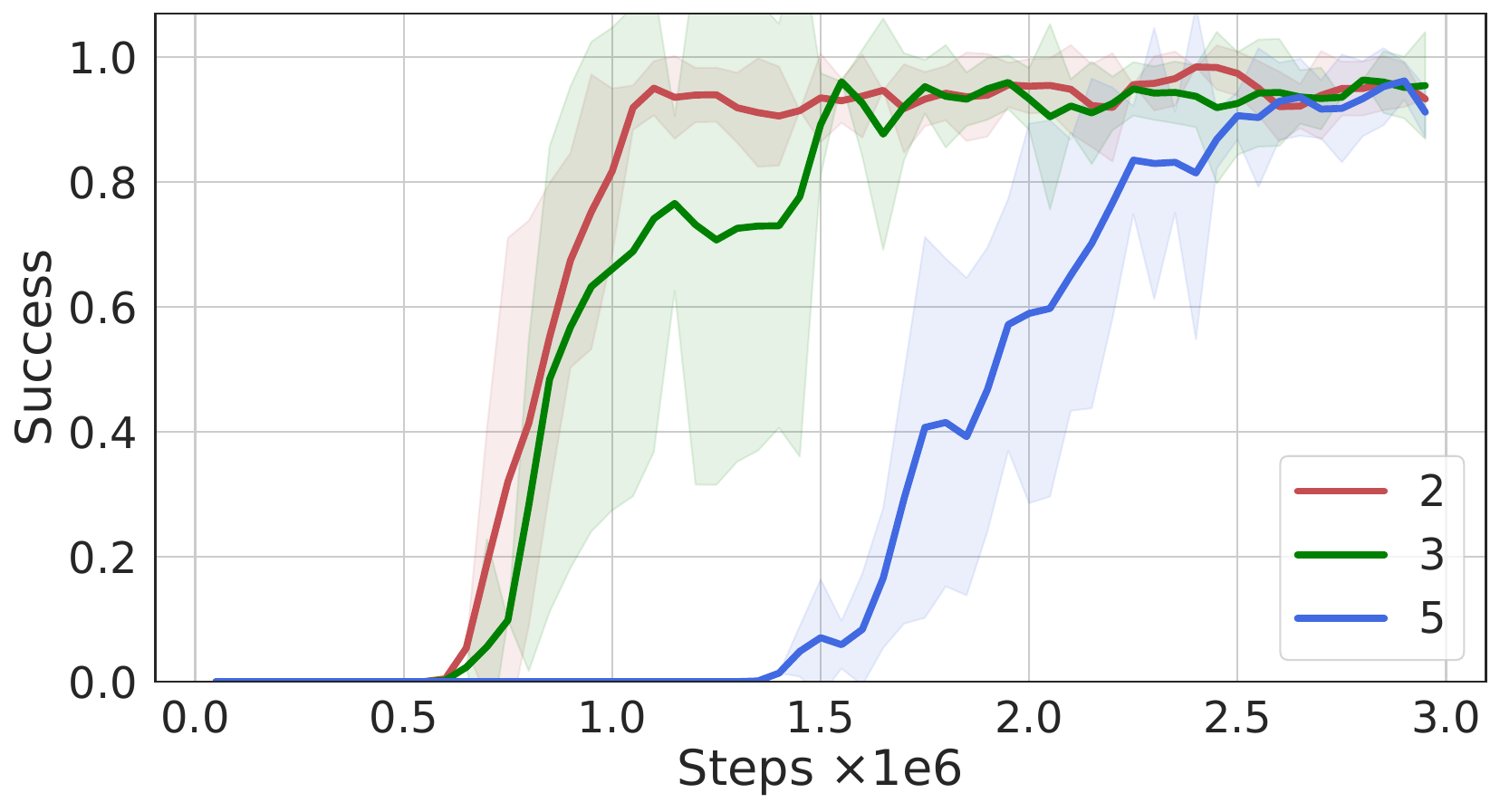}
		\end{minipage}
	}%
 
	\subfigure[Scaling Factor $\beta$, Power $n$]{
		\begin{minipage}[t]{0.47\linewidth}
			\centering
			\includegraphics[width=\linewidth]{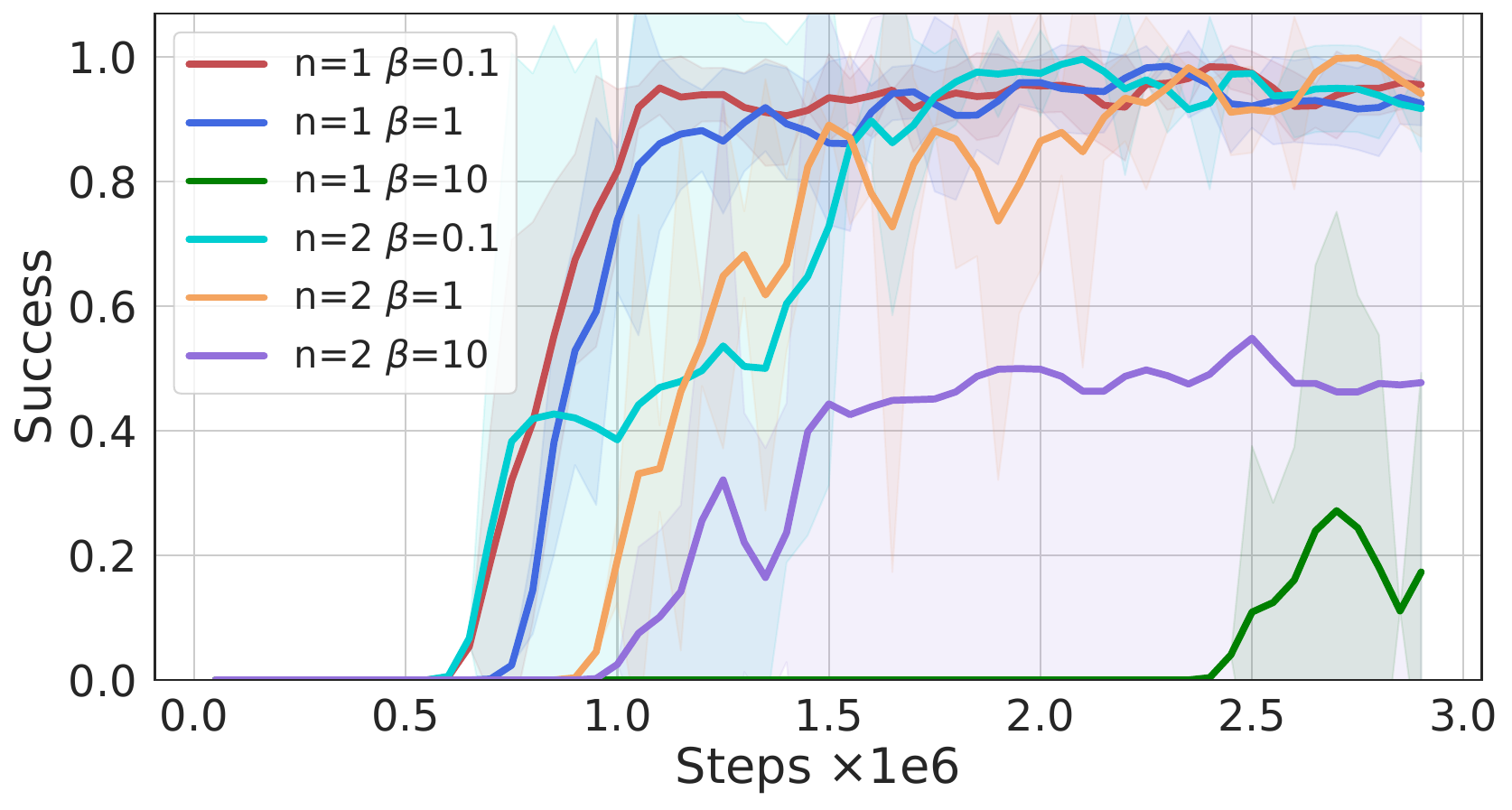}
		\end{minipage}
	}%
    \subfigure[Subgoal Selection Interval $c$]{
		\begin{minipage}[t]{0.47\linewidth}
			\centering
			\includegraphics[width=\linewidth]{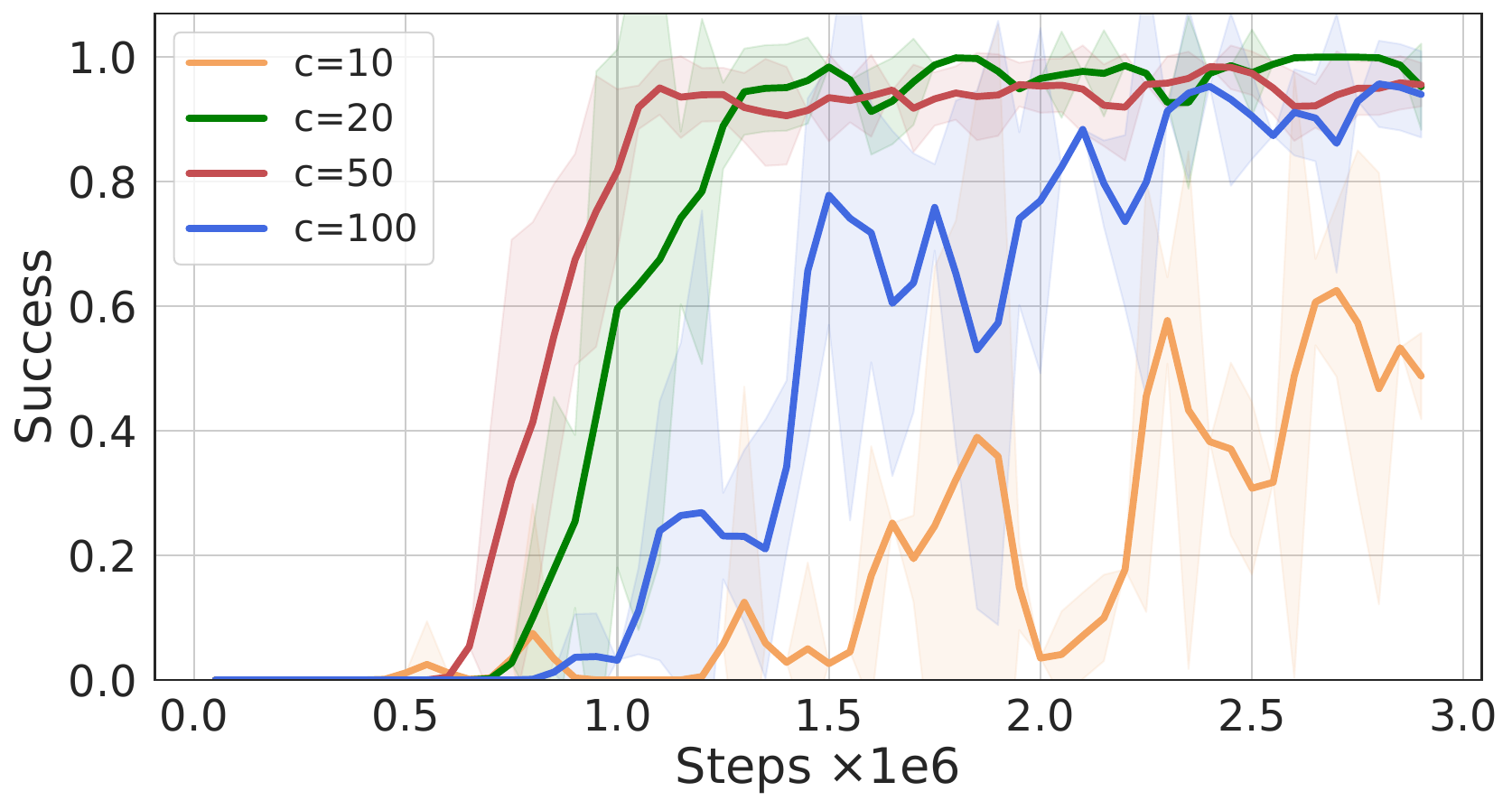}
		\end{minipage}
	}%
 
	\subfigure[Landmark Sampling Strategy]{
		\begin{minipage}[t]{0.47\linewidth}
			\centering
			\includegraphics[width=\linewidth]{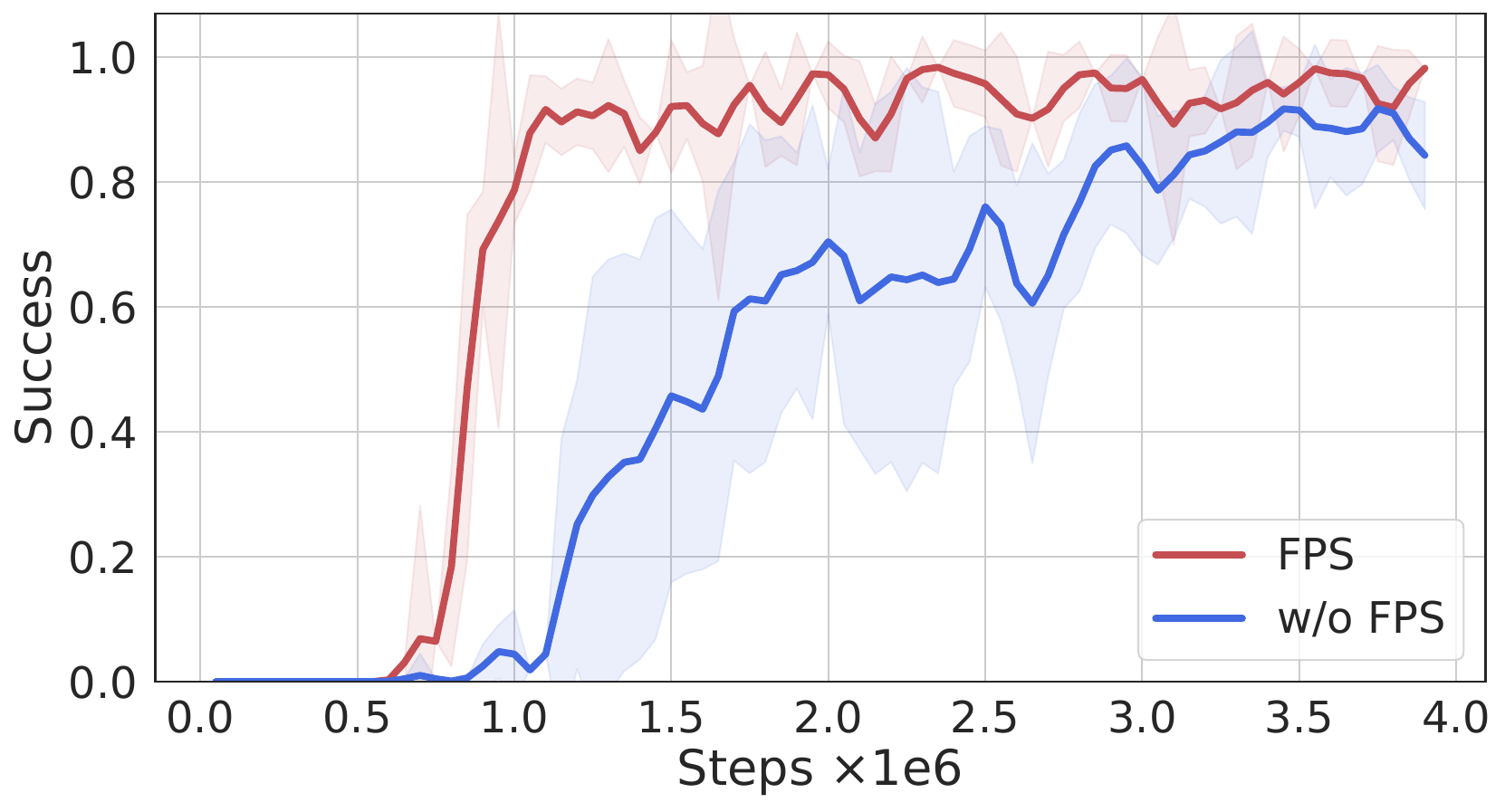}
		\end{minipage}
	}
	\subfigure[Landmark Number]{
		\begin{minipage}[t]{0.47\linewidth}
			\centering
			\includegraphics[width=\linewidth]{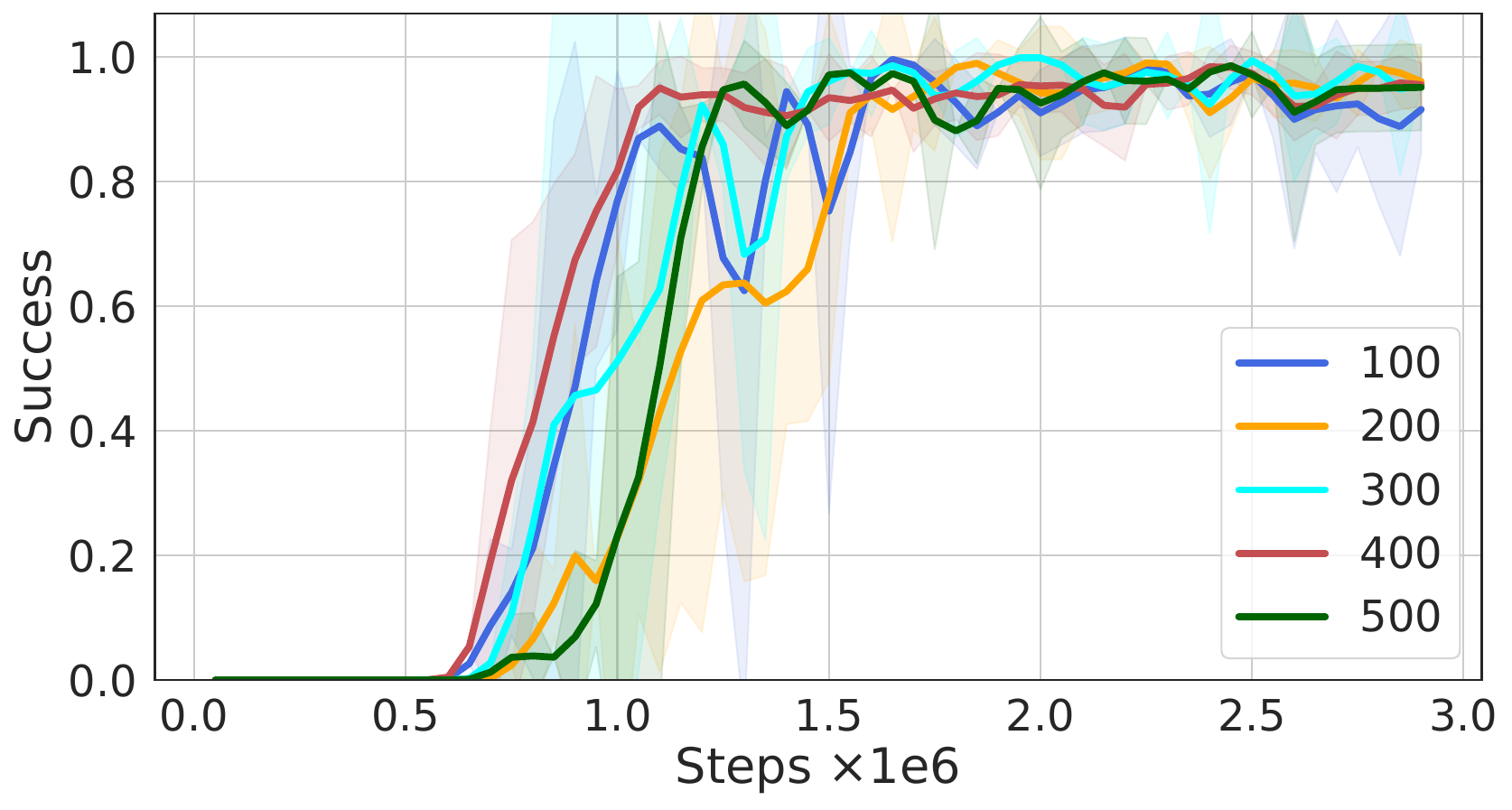}
		\end{minipage}
	}%

	\centering
	\caption{Ablation studies of hyperparameter selection in the Ant Maze task.}\vspace{-10pt}
	\label{fig:ablation2}
\end{figure}

\subsection{Comparative Experiments}
We compare HILL with the SOTA baselines, which include: (1) \textit{HESS} \cite{li2021active}: a GCHRL method realizing active exploration based on stable subgoal representation learning. (2) \textit{LESSON} \cite{li2020learning}: a GCHRL method learning subgoal representations with slow dynamics. 
(3) \textit{HIRO} \cite{nachum2018data}: 
a GCHRL method using off-policy correction for efficient exploitation. (4) \textit{SAC} \cite{haarnoja2018soft}: the base RL method we used for training the bi-level policies. Note that we use the original implementations of LESSON and HESS on all tasks. 

The experimental results, as presented in Figure~\ref{main results}, demonstrate that HILL outperforms all baselines in terms of both sample efficiency and asymptotic performance.
This success can be attributed to the use of efficient subgoal representations with temporal coherence and a subgoal selection strategy that effectively balances exploration and exploitation.
In comparison, HESS underperforms our method due to its focus on exploring novel and reachable subgoals, lacking consideration of the potential benefits that subgoals can provide in completing the source task.
LESSON yields an uptrend similar to HILL in Ant Push in the early stage. However, its unstable subgoal representations lead to a subsequent drop in performance.
HIRO improves sample efficiency by relabeling subgoals to maximize the likelihood of past low-level action sequences. However, it lacks active exploration and fails to achieve a balance between exploration and exploitation.
SAC performs poorly across all tasks, showing hierarchical advantages in solving long-term tasks with sparse rewards.

\subsection{Visualization Analysis}

\subsubsection{Exploration and Exploitation}
We visualize the exploration and exploitation statuses by counting the visits of states in the state space and representations in the latent space, respectively, as shown in Figure~\ref{fig:exploration}.
With the guidance of the novelty measure, HILL actively selects subgoals that are less explored, resulting in an extensive range of representations in the latent space and a dispersed distribution in the state space. 
As training progresses, the cumulative visits to various latent representations become uniform, and the utility measure enables more precise estimates and converges gradually.
Consequently, episodes in the state and latent spaces gradually stabilize and exhibit little difference. The active exploration in the early stage and adequate exploitation in the later stage demonstrate the effectiveness of our proposed balanced strategy.

\subsubsection{Representation Learning}
We run $10$ episodes under the most challenging setting of the Ant Maze task and visualize their subgoal representation learning processes. We use a color gradient from purple to yellow to indicate the trend of an episode, as shown in Figure \ref{fig:representation}, where purple represents the start position, and yellow represents the end position. HILL learns representations with temporal coherence based on environmental transition relationships in an early stage. Adjacency states along an episode are clustered to similar latent representations, and ones multiple steps apart are pushed away in the latent space. These demonstrate the effectiveness and efficiency of the proposed negative-power contrastive representation learning objective.

\subsection{Ablation Study on Various Components}
We conduct ablation studies on various components by incrementally adding critical modules to the basic GCHRL framework. The learning curves are shown in Figure~\ref{fig:ablation1}. 
Basic GCHRL, which solely relies on SAC as the base RL optimizer of both levels, fails to learn policies efficiently due to the large exploration space and sparse rewards. The non-stationarity introduced by joint training hinders bi-level policy updates.
Our proposed contrastive representation learning objective compresses the exploration space by considering the temporal coherence in GCHRL (B+CSRL).
The use of HER alleviates non-stationary issues and enables low-level policies to converge faster, resulting in better bi-level cooperation, thus improving exploration efficiency (B+CSRL+HER).
Building latent landmark graphs and measuring novelties at nodes help agents actively explore novel subgoals (B+CSRL+HER+G-N).
Finally, HILL (B+CSRL+HER+G-N-U) is obtained by considering both the novelty and utility measures on latent landmark graphs. By selecting the most valuable subgoal that balances the two measures, HILL realizes effective exploration-exploitation trade-offs, resulting in the fastest convergence speed and highest asymptotic performance.

\subsection{Ablation Study on HyperParameters}
We set up ablation studies on several typical hyperparameters of HILL in the Ant Maze task. Figure \ref{fig:ablation2} shows the evaluation results. We conclude that our method is robust over an extensive range of hyperparameters. 
\subsubsection{Stable ratio $k$} Sampling the top $k$ of triplets with lower representation losses helps to stably constrain changes of well-learned representations. However, a large $k$ may disturb the update of subgoal representation learning, while a small $k$ reduces the effectiveness of stability regularization, as indicated in Figure~\ref{fig:ablation2} (a). We set $k = 0.3$ for all tasks.
\subsubsection{Representation Dimension $d$} 
$d$ determines the degree of information compression. The smaller $d$ is, the more compact the representation abstracted by $\phi$ is.  
The results in Figure \ref{fig:ablation2} (b) indicate that HILL achieves better performance when $d=2$. Therefore, we set $d$ to $2$ for all tasks.
\subsubsection{Scaling Factor $\beta$, Power $n$} 
$\beta$ and $n$ both affect the performance of subgoal representation learning. A smaller $n$ or a larger $\beta$ pushes negative instances further away. As shown in Figure \ref{fig:ablation2} (c), HILL maintains robustness when $\beta$ is small and achieves better performance when $n=1,\beta=0.1$. Therefore, we set $n=1,\beta=0.1$ for all tasks.
\subsubsection{Subgoal Selection Interval $c$}
$c$ affects the subgoal selection interval and the learning difficulty of the low-level policy. A smaller $c$ results in faster convergence of the low-level policy but makes high-level decision-making more challenging. This adversarial relationship is verified in Figure \ref{fig:ablation2} (d). We set $c=50$ for all tasks and baselines in our experiments.
\subsubsection{Landmark Sampling Strategy}
Figure~\ref{fig:ablation2} (e) shows that HILL converges faster with FPS than with uniform sampling. This is due to FPS's ability to identify representations that maximize the coverage of the latent space, which facilitates the agent's exploration of new regions. Therefore, FPS is adopted as the landmark sampling strategy for all tasks.
\subsubsection{Landmark Number}
The number of landmarks has a significant impact on the coverage of the latent space and the transition distance between nearby landmarks. Inadequate landmark numbers lead to poor coverage, while excessive numbers increase the burden of value estimation. The results in Figure \ref{fig:ablation2} (f) indicate that HILL achieves better performance when the number is $400$. Therefore, we use it as the landmark number for all tasks.

\section{Conclusion}
This work proposes to address the exploration and exploitation dilemma in hierarchical reinforcement learning by dynamically constructing latent landmark graphs (HILL).
We propose a contrastive representation learning objective to learn subgoal representations that comply with the temporal coherence in GCHRL, as well as a latent landmark graph structure that balances subgoal selection for effective exploration-exploitation trade-offs.
Empirical results demonstrate that HILL significantly outperforms the SOTA GCHRL methods on numerous continuous control tasks with sparse rewards. Visualization analyses and ablation studies further highlight the effectiveness and efficiency of various HILL components.

\bibliographystyle{IEEEtranN}
\bibliography{aaai23_sim}

\end{document}